\documentclass[runningheads]{llncs}
\usepackage{graphicx}
\usepackage[tight,footnotesize]{subfigure}
\usepackage[cmex10]{amsmath}
\usepackage{amsfonts}

\begin{document}
\title{Neural Style Difference Transfer and Its Application to Font Generation}
%
%
\author{Gantugs Atarsaikhan \and
Brian Kenji Iwana\and
Seiichi Uchida}
%
\authorrunning{G. Atarsaikhan et al.}
%
\institute{Kyushu University, Fukuoka, Japan \\
\email{\{gantugs.atarsaikhan, brian, uchida\}@human.ait.kyushu-u.ac.jp}}
\maketitle              
\begin{abstract}
Designing fonts requires a great deal of time and effort. It requires professional skills, such as sketching, vectorizing, and image editing. Additionally, each letter has to be designed individually. In this paper, we introduce a method to create fonts automatically. In our proposed method, the difference of font styles between two different fonts is transferred to another font using neural style transfer. Neural style transfer is a method of stylizing the contents of an image with the styles of another image. We proposed a novel neural style difference and content difference loss for the neural style transfer. With these losses, new fonts can be generated by adding or removing font styles from a font. We provided experimental results with various combinations of input fonts and discussed limitations and future development for the proposed method.

\end{abstract}
\section{Introduction}
\label{sec:intro}

Digital font designing is a highly time-consuming task. It requires professional skills, such as sketching ideas on paper and drawing with complicated software. Individual characters or letters has many attributes to design, such as line width, angles, stripes, serif, and more. Moreover, a designer has to design all letters character-by-character, in addition to any special characters. For example, the Japanese writing system has thousands of Japanese characters that needs to be designed individually. Therefore, it is beneficial to create a method of designing fonts automatically for people who have no experience in designing fonts. It is also beneficial to create a way to assist font designers by automatically drawing fonts.

\begin{figure}[!t]
    \centering
    \includegraphics[width=\columnwidth]{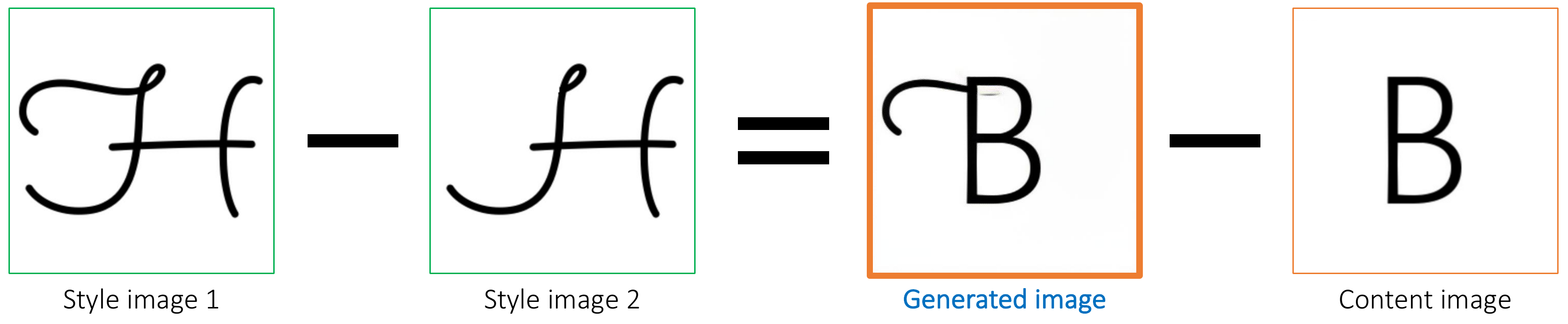}
    \caption{An example results of the proposed method. Style difference between the style image 1 and 2 is transferred onto the content image by equalling to the style difference between the newly generated image and the content image. }
    \label{fig:proposed_example}
\end{figure}

On the other hand, there are a large number of fonts that have already designed. Many of them have different font styles, such as \textbf{bold}, \textit{italic}, \textsc{serif} and \textsc{\textsf{sans serif}}. There are many works done to create new fonts by using already designed fonts~\cite{uchida2015exploring,miyazaki2017automatic}. In this paper, we chose an approach to find the difference between two fonts and transfer it onto a third font in order to create a new font. For example, using a font with serifs and a font without serifs to transfer the serif difference to a third font that originally lacked serifs, as shown in Fig.~\ref{fig:proposed_example}.

\begin{figure}[!t]
    \centering
    \includegraphics[width=\columnwidth]{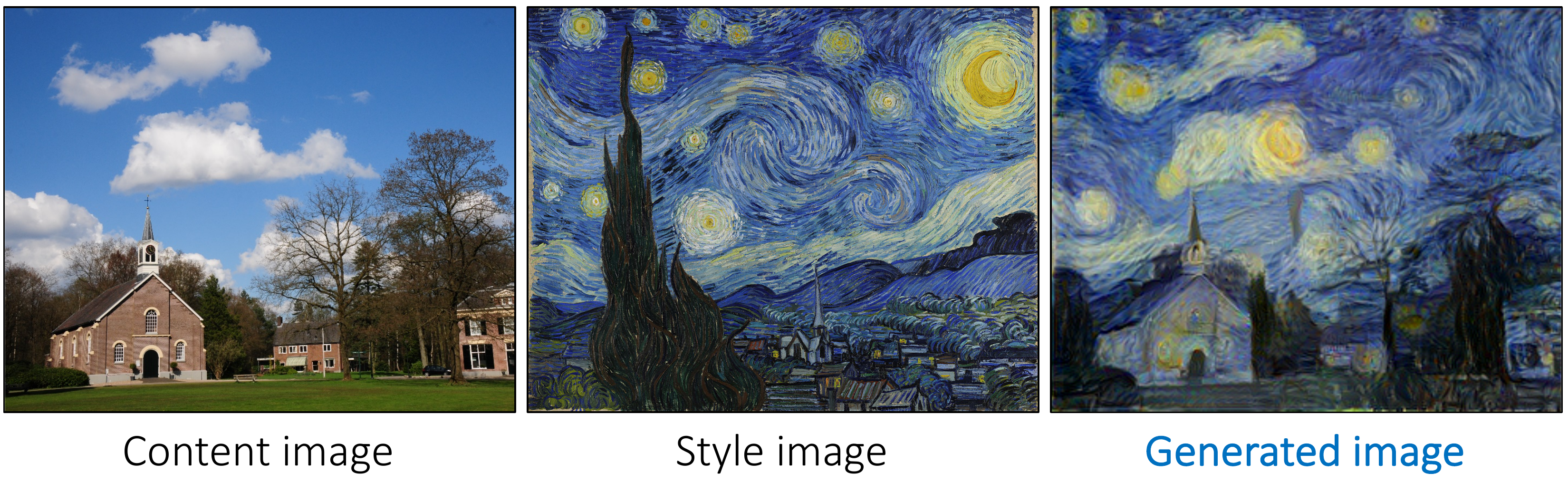}
    \caption{An example of the NST. Features of the style image are blended into the structure of the content image in the generated result image.}
    \label{fig:nst_example}
\end{figure}

In recent years, the style transfer field has progressed rapidly with the help of Convolutional Neural Networks~(CNN)~\cite{lecun1998gradient}.
Gatys et al.~\cite{Gatys2016styletransfer} first used a CNN to synthesize an image using the style of an image and the content of another image using Neural Style Transfer~(NST). In the NST, the content is regarded as feature maps of a CNN and the style is determined by the correlation of feature maps in a Gram matrix. The Gram matrix calculates how correlated the feature maps are to each other. An example of the NST is shown in Fig.~\ref{fig:nst_example}. The content image and style image are mixed by using features from the CNN to create a newly generated image that has contents (buildings, sky, trees, etc.) from the content image and style (swirls, paint strokes, etc.) from the style image. There are also other style transfer methods, such as a ConvDeconv network for real-time style transfer~\cite{Johnson} and methods that utilize Generative Adversarial Networks~(GAN)~\cite{Pix2Pix}.

The purpose of this paper is to explore and propose a new method to create novel fonts automatically. Using NST, the contents and styles of two different fonts have been found and their difference is transferred to another font to generate a new synthesized font. Fig.~\ref{fig:proposed_example} shows an example results of our proposed method. We provided experimental results and inspected the performance of our method with various combinations of content and style images. 

The main contributions of this paper are as follows.\\
\begin{enumerate}
    \item This is the first attempt and trial on transferring the difference between neural styles onto the content image.
    \item Proposed a new method to generate fonts automatically to assist font designers or non-professionals.
\end{enumerate}

The remaining of this paper is arranged as follows. In Section~\ref{sec:related_work}, we discuss previous works on font generation, style transfer fields, and font generation using CNN. The proposed method is explained in Section~\ref{sec:method} in detail. Then, Section~\ref{sec:experimental_results} examines the experiments and the results. Lastly, we conclude in Section~\ref{sec:conclusion} with concluding remarks and discussions for improvements.

\section{Related Work}
\label{sec:related_work}

\subsection{Font Generation}
Various attempts have been made to create fonts automatically. One approach is to generate font using example fonts. Devroye and McDougall~\cite{devroye1995random} created handwritten fonts from handwritten examples. Tenenbaum and Freeman~\cite{tenenbaum2000separating} clustered fonts by its font styles and generated new fonts by mixing font styles with other fonts. Suveeranont and Igarashi~\cite{suveeranont2009feature,suveeranont2010example} generated new fonts from user-defined examples. Tsuchiya et al.~\cite{Tsuchiya2014} also used example fonts to determine features for new fonts. Miyazaki et al.~\cite{miyazaki2017automatic} extracted strokes from fonts and generated typographic fonts. 

Another approach in generating fonts is to use transformations or interpolations of fonts. Wada and Hagiwara~\cite{wada2003japanese} created new fonts by modifying some attributes of fonts, such as slope angle, thickness, and corner angle. Wand et al.~\cite{wang2008style} transformed strokes of Chinese characters to generate more characters. Campbell and Kautz~\cite{campbell2014learning} created new fonts by mapping non-linear interpolation between multiple existing fonts. Uchida et al.~\cite{uchida2015exploring} generated new fonts by finding fonts that are simultaneously similar to existing fonts.

Lake et al.~\cite{lake2015human} generated handwritten fonts by capturing example patterns with the Bayesian program learning method. Baluja et al.~\cite{baluja2017learning} learned styles from four characters to generate other characters using CNN-like architecture. Recently, many studies use machine learning for font design. Atarsaikhan et al.~\cite{tugs} used the NST to synthesize a style font and a content font to generate a new font. Also, GANs have been used to generate new fonts~\cite{abe2017font,Lyu2017AutoEncoderGG,Sun2018PyramidEG}. Lastly, fonts have been stylized with patch-based statistical methods~\cite{yang2017awesome}, NST~\cite{atarsaikhan2018contained} and GAN methods~\cite{Azadi_2018_CVPR,zhao2018webfontgan,DBLP:journals/corr/abs-1812-06384}.

\subsection{Style Transfer}
The first example-based style transfer method was introduced in "Image Analogies" by Hertzmann et al.~\cite{hertzmann2001imageanalogies}. Recently, Gatys et al. developed the NST~\cite{Gatys2016styletransfer} by utilizing a CNN. There are two types of NST methods, i.e. image optimization-based and network optimization-based. NST is the most popular image optimization-based method. However, the original NST~\cite{Gatys2016styletransfer} was introduced for artistic style transfer and works have been done for photorealistic style transfer~\cite{luan2017deep,mechrez2017Photorealistic}. New losses are introduced for stable results and preserving low-level features~\cite{risser2017stable,li2017laplacianloss}. Additionally, improvements, such as semantic aware style transfer~
\cite{mechrez2018contextual,champandard2016semantic},
controlled content features~\cite{gatys2016preserving,gatys2016controlling} were proposed.
Also, Mechrez et al.~\cite{mechrez2018contextual} and Yin et al.~\cite{yin2016content}, achieved semantic aware style transfer.

In network optimization-based style transfer, first, neural networks are trained for a specific style image and then this trained network is used to stylize a content image. A ConvDeconv neural network~\cite{Johnson} and a generative network ~\cite{ulyanov2016texture} are trained for style transfer. They improved for photorealistic style transfer~\cite{li2018closed} and semantic style transfer~\cite{champandard2016semantic,chen2016towards}.

There are many applications of the NST due to its non-heuristic style transfer qualities. It has been used for video style transfer~\cite{anderson2016deepmovie,joshi2017bringing}, portrait~\cite{selim2016painting}, fashion~\cite{Jiang_2017}, and creating doodles~\cite{champandard2016semantic}. Also, character stylizing techniques have been proposed by improving the NST~\cite{atarsaikhan2018contained} or using NST as part of the bigger network~\cite{Azadi_2018_CVPR}.

\section{Neural Style Difference Transfer for Font Generation}
\label{sec:method}

\begin{figure}[!t]
\centering
\includegraphics[width=\columnwidth]{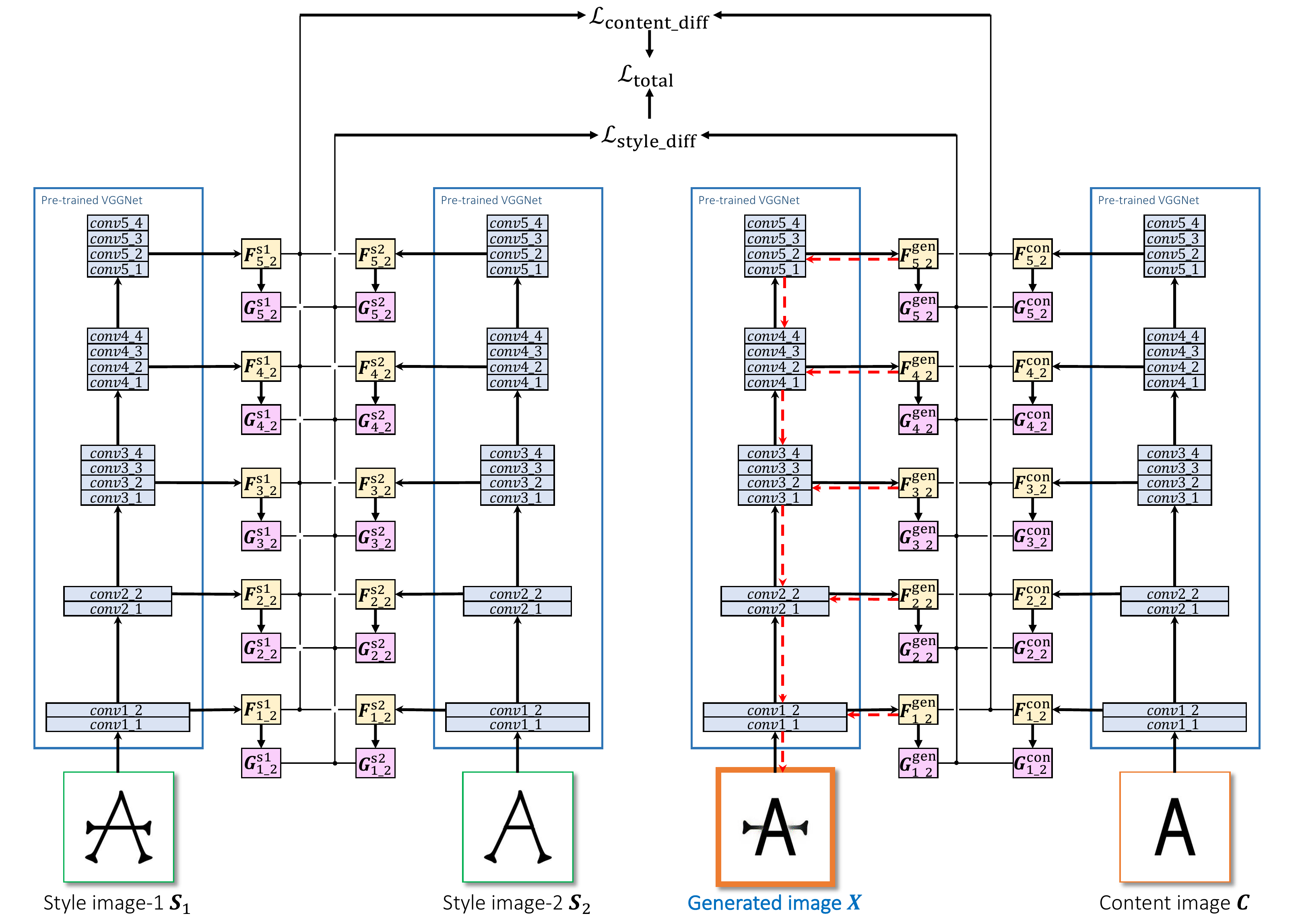}
\caption{The overview font generation with neural style difference transfer. Yellow blocks show feature maps from one layer and purple blocks show Gram matrices calculated using the feature maps. }
\label{fig:whole_process}
\end{figure}

The overview of the proposed method is shown in Fig.~\ref{fig:whole_process}. A pre-trained CNN called Visual Geometry Group~(VGGNet)~\cite{Simonyan} is used to input the images and extract their feature maps on various layers. VGGNet is trained for natural scene object recognition, thus making it extremely useful for capturing various features from various images. The feature maps from higher layers show global arrangements of the input image and the feature maps from lower layers express fine details of the input image. Feature maps from specified content layers are regarded as content representations and correlations of feature maps on specified style layers are regarded as style representations.

There are three input images: style image-1 $\pmb{S}_1$, style image-2 $\pmb{S}_2$ and the content image $\pmb{C}$. $\pmb{X}$ is the generated image that is initiated as the content image or as a random image. 
The difference of content representation and style representation between style images are transferred onto the content image by optimizing the generated image.
First, the content difference and style difference of style images are calculated and stored. Also, the differences are calculated between the generated image and the content image. The content difference loss is calculated as the sum of the layer-wise mean squared errors between the differences of the features maps in the content representation. The style difference loss is calculated in the same way but between the differences of the Gram matrices in the style representation. Then, the content difference loss and style difference loss are accumulated into the total loss. Lastly, the generated image is optimized through back-propagation to minimize the total loss. By repeatedly optimizing the generated image with this method, the style difference between the style images is transferred to the content image.

\subsection{Neural Style Transfer}
\label{sec:nst}

Before explaining the proposed method in detail, we will briefly discuss NST. In the NST, there are two inputs: a content image $\pmb{C}$ and a style image $\pmb{S}$. The image to be optimized is the generated image $\pmb{X}$. It also uses a CNN, i.e. VGGNet to capture features of the input images and create a Gram matrix for the style representation from the captured feature maps. A Gram matrix is shown in Eq.~\ref{eq:gramm}, where $\pmb{D}_{l}$ is a matrix that consists of flattened feature maps of layer $l$ as shown in Eq.~\ref{eq:dl}. The Gram matrix calculates the correlation value of feature maps from one layer to each other and stores it into a matrix.

\begin{equation}
\label{eq:gramm}
     \pmb{G}_{l} = \pmb{D}_l(\pmb{D}_l)^\top,
\end{equation}
where, 
\begin{equation}
\label{eq:dl}
    \pmb{D}_l = \{\Vec{\pmb{F}}_1,..., \Vec{\pmb{F}}_{n_l},...,\Vec{\pmb{F}}_{N_l}\}.
\end{equation}

First, the style image $\pmb{S}$ is input to the VGGNet. Its feature maps $\mathbb{F}^\mathrm{style}$ on given style layers are extracted
and their Gram matrices $\mathbb{G}^\mathrm{style}$ are calculated. Next, the content image $\pmb{C}$ input to the VGGNet and its feature maps $\mathbb{F}^\mathrm{content}$ on given content layers are extracted and stored. Lastly, the generated image $\pmb{X}$ is input to the network. Its Gram matrices $\mathbb{G}^\mathrm{generated}$ on style layers and feature maps $\mathbb{F}^\mathrm{generated}$ from content layers are found. 

Then, by using the feature maps and Gram matrices, the content loss and style loss are calculated as, 
\begin{equation}
\label{eq:content_loss_nst}
    \mathcal{L}_\mathrm{content} = \sum^{L_c}_{l}\frac{w^\mathrm{content}_l}{2N_lM_l}\sum^{N_l}_{n_l}\sum^{M_l}_{m_l}(F^{\mathrm{generated}}_{n_l,m_l}-F^{\mathrm{content}}_{n_l,m_l})^2,
\end{equation}
and
\begin{equation}
    \label{eq:style_loss_nst}
    \mathcal{L}_\mathrm{style} = \sum^{L_s}_l\frac{w^\mathrm{style}_l}{4N^2_lM^2_l}\sum^{N_l}_i\sum^{N_l}_j(G^\mathrm{generated}_{l,i,j}-G^\mathrm{style}_{l,i,j})^2,
\end{equation}
where $L_c$ and $L_s$ are the number of layers, $N_l$ is the number of feature maps, $M_l$ is the number of elements in one feature map, $w^\mathrm{content}_l$ and $w^\mathrm{style}_l$ are weights for layer $l$. Lastly, the content loss $\mathcal{L}_\mathrm{content}$ and the style loss $\mathcal{L}_\mathrm{style}$ are accumulated into the total loss $\mathcal{L}_\mathrm{total}$ with weighting factors $\alpha$ and $\beta$:
\begin{equation}
\label{eq:total_loss_nst}
	\mathcal{L}_\mathrm{total} = \alpha \mathcal{L}_\mathrm{content} + \beta \mathcal{L}_\mathrm{style}.
\end{equation}
Once the total loss $\mathcal{L}_\mathrm{total}$ is calculated, the gradients of content layers, style layers, and generated image $\pmb{X}$ are determined by back-propagation. Then, to minimize the total loss $\mathcal{L}_\mathrm{total}$, only the generated image $\pmb{X}$ is optimized. By repeating these steps multiple times, the style from the style image are transferred to the content image in the form of a generated image.

In the NST, the goal of the optimization process is to match the styles of the generated image to those of the style image, and feature maps of the generated image to those of the content image. However, in the proposed method, the goal of the optimization process is to match the style differences between the generated image and the content image to those of style images, as well as, differences of content difference between the generated image and the content image to those of style images.

\subsection{Style Difference Loss}
\label{sec:style_loss}

Let $\pmb{G}^\mathrm{style1}_{l}$ and $\pmb{G}^\mathrm{style2}_{l}$ be the Gram matrices of feature maps on layer $l$, when style image-1 $\pmb{S}_1$ and style image-2 $\pmb{S}_2$ are input respectively. Then, the style difference between the style images on layer $l$ is defined as,
\begin{equation}
    \label{eq:style_diff_style}
    {\Delta}\pmb{G}^\mathrm{style}_l = \pmb{G}^\mathrm{style1}_l-\pmb{G}^\mathrm{style2}_l,
\end{equation}
Similarly, the style difference between the generated image $\pmb{X}$ and the content image $\pmb{C}$ is defined as,
\begin{equation}
    \label{eq:style_diff_generated}
    {\Delta}\pmb{G}^\mathrm{generated}_l = \pmb{G}^\mathrm{generated}_l-\pmb{G}^\mathrm{content}_l.
\end{equation}
Consequently, the style loss is the mean squared error between the style differences:
\begin{equation}
    \label{eq:style_loss}
    \mathcal{L}_\mathrm{style\_diff} = \sum^L_l\frac{w^\mathrm{style}_l}{4N^2_lM^2_l}\sum^{N_l}_i\sum^{N_l}_j({\Delta}G^\mathrm{generated}_{l,i,j}-{\Delta}G^\mathrm{style}_{l,i,j})^2,
\end{equation}
where $w^\mathrm{style}_l$ is a weighting factor for an individual layer $l$. Note, it can be set to zero to ignore a specific layer. The style difference loss in the proposed method means that the difference of correlations of feature maps (Gram matrix) of the generated and content images~($\pmb{X}$ and $\pmb{C}$) are forced to match that of style images~($\pmb{S}_1$ and $\pmb{S}_2$) through optimization of the generated image, so that the style difference (e.g. bold and light fonts styles, italic or regular fonts styles) are transferred. 

\subsection{Content Difference Loss}
\label{sec:content_loss}

Extracting the feature maps on a layer $l$ as $\pmb{F}^{\mathrm{style1}}_{l}$ and $\pmb{F}^{\mathrm{style2}}_{l}$ for the style images, the content difference between the style images on layer $l$ are defined as follows,
\begin{equation}
\label{eq:content_diff_style}
    {\Delta}\pmb{F}^\mathrm{style}_l = \pmb{F}^{\mathrm{style1}}_l - \pmb{F}^{\mathrm{style2}}_l . 
\end{equation}
Using the same rule, the content difference between the generated and content images on layer $l$ is defined as,
\begin{equation}
\label{eq:content_diff_generated}
    {\Delta}\pmb{F}^\mathrm{generated}_l = \pmb{F}^{\mathrm{generated}}_l - \pmb{F}^{\mathrm{content}}_l , 
\end{equation}
where $\pmb{F}^{\mathrm{generated}}_{l}$ is the feature maps on layer $l$ when the generated image $\pmb{X}$ is input, and $\pmb{F}^{\mathrm{content}}_{l}$ is the feature maps when the content image $\pmb{C}$ is input. By using content differences of the two style images and the generated and content images, the content difference loss is calculated as,
\begin{equation}
\label{eq:content_loss}
    \mathcal{L}_\mathrm{content\_diff} = \sum^{L}_{l}\frac{w^\mathrm{content}_l}{2N_lM_l}\sum^{N_l}_{n_l}\sum^{M_l}_{m_l}({\Delta}F^{\mathrm{generated}}_{l,n_l,m_l}-{\Delta}F^{\mathrm{content}}_{l,n_l,m_l})^2,
\end{equation}
where $w^\mathrm{content}_l$ is weighting factor for layer $l$. Layers also can be ignored by setting $w^\mathrm{content}_l$ to zero. The content difference loss captures the difference in global feature of the style images, e.g, difference in serifs.

\subsection{Style Transfer}
\label{synthesizing}

For style transfer, a generated image $\pmb{X}$ is optimized to simultaneously match the style difference on style layers and the content difference on content layers. Thus, a loss function is created and minimized: The total loss $\mathcal{L}_\mathrm{total}$ which is the accumulation of the content difference loss $\mathcal{L}_\mathrm{content\_diff}$ and the style difference loss $\mathcal{L}_\mathrm{style\_diff}$ written as,
\begin{equation}
\label{eq:total_loss}
    \mathcal{L}_\mathrm{total} = \mathcal{L}_\mathrm{content\_diff} + \mathcal{L}_\mathrm{style\_diff}.
\end{equation}
With the total loss, the gradients of pixels on the generated image are calculated using back-propagation and used for an optimization method, such as L-BFGS or Adam. We found from experience that L-BFGS requires fewer iterations and produces better results. Also, we use the same image size for each input image for comparable feature sizes. Due to plain subtracting operation in Eq.~\ref{eq:style_diff_style}, Eq.\ref{eq:style_diff_generated}, Eq.\ref{eq:content_diff_style}, and Eq.\ref{eq:content_diff_generated}, input images have to be chosen carefully. Font styles of $\pmb{S}_2$ must be similar to that of $\pmb{C}$. Because of the subtraction process, the generated image $\pmb{X}$ is most likely optimized to have similar font styles with style image-1 $\pmb{S}_1$. Moreover, contrary to the NST, we do not use weighting factors for content loss and style loss, instead, individual layers are weighted.

\begin{figure}[!t]
    \centering
    \includegraphics[width=\columnwidth]{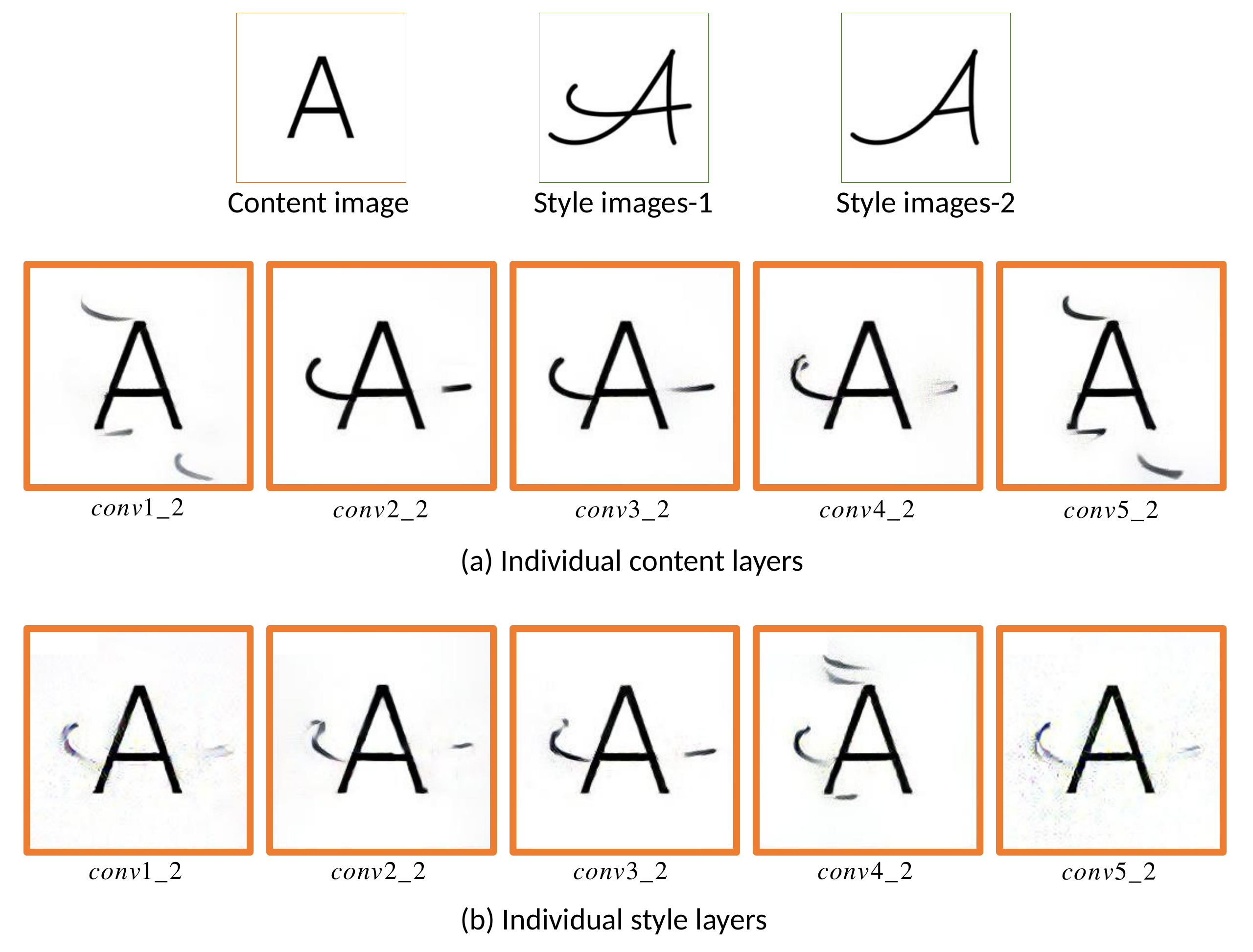}
    \caption{Various weights for content and style layers.}
    \label{fig:weights}
\end{figure}

\begin{figure}[!t]
    \centering
    \includegraphics[width=\columnwidth, trim= 0 720 0 0, clip]{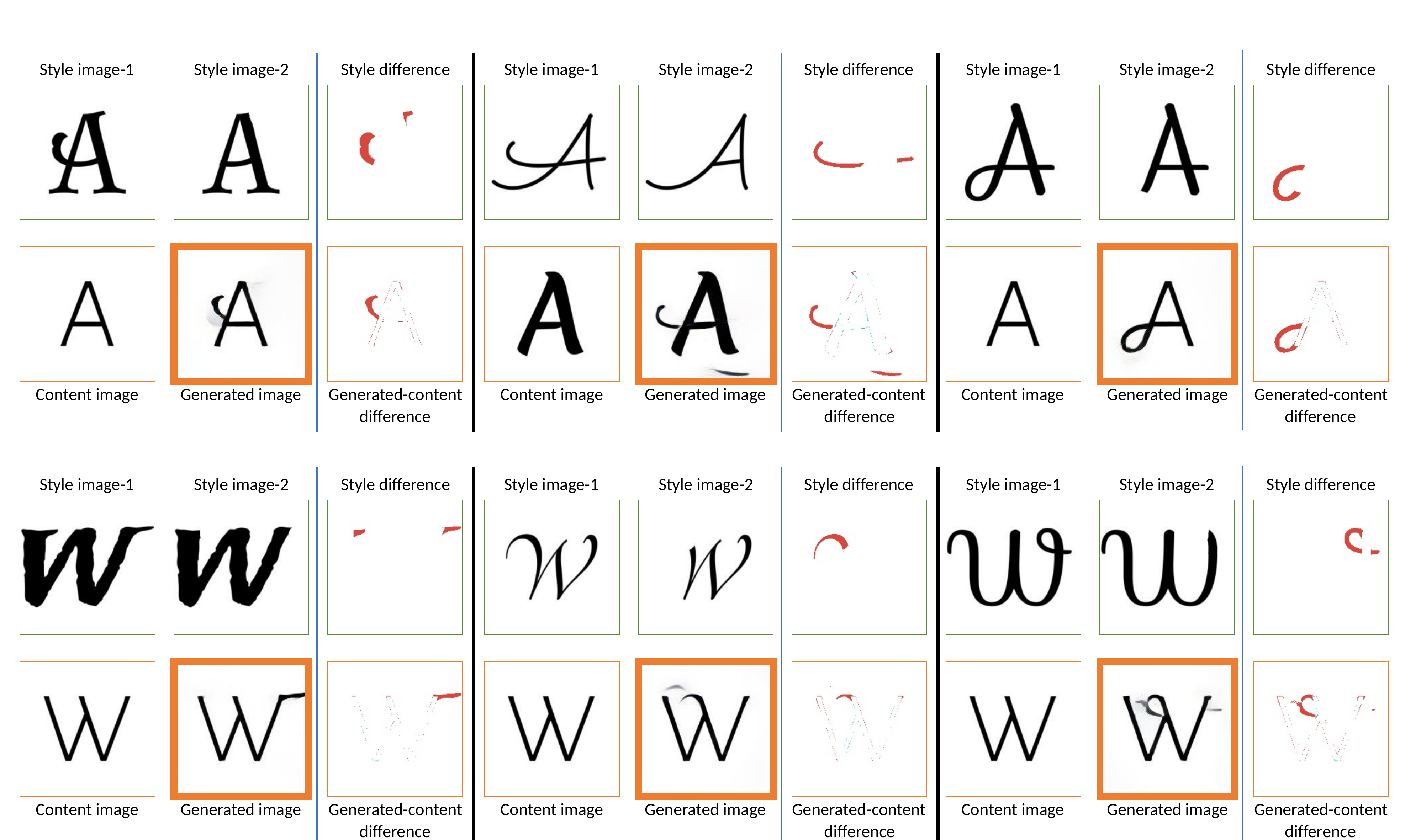}
    \caption{Results of transferring missing parts to another font. In the style difference image, red shows parts that exist in the style image-1 but does not exist in the style image-2. Blue shows the reverse.  Moreover, parts that are transferred onto the content image are visualized in red, and parts that are erased from the content image are shown blue in generated-content difference images. As a clarification, style difference and generated-content difference are mere visualizations from the input images and generated image, they are not input or results images themselves.}
    \label{fig:missing_part}
\end{figure}

\section{Experimental Results}
\label{sec:experimental_results}

In the experiments below except specified, feature maps for the style difference loss are taken from the style layers, $conv1\_2$, $conv2\_2$, $conv3\_2$, $conv4\_2$, $conv5\_2$ with style weights $\pmb{w}^\mathrm{style}=\{\frac{10^3}{64^2},\frac{10^3}{128^2},\frac{10^3}{256^2},\frac{10^3}{512^2},\frac{10^3}{512^2}\}$, and feature maps for the content difference loss is taken from content layer $conv4\_2$ with content weight $w_{conv4\_2}=10^4$ on VGGNet. Generated image $\pmb{X}$ is initialized with the content image $\pmb{C}$. The optimization process is stopped at 1,000th step with more than enough convergence. Also, due to black pixels having a zero value, input images are inverted before inputting to the VGGNet and inverted back for the visualization. 

\subsection{Content and Style Layers}
\label{sec:layers}

Fig.~\ref{fig:weights} shows results using various content and style layers individually. We used a sans-serif font for the content image, and tried to transfer the horizontal line style difference between style fonts. In Fig.~\ref{fig:weights}a, we experimented on using each content layers while the weights for the style layers are fixed as $\pmb{w}^\mathrm{style}=\{\frac{10^3}{64^2},\frac{10^3}{128^2},\frac{10^3}{256^2},\frac{10^3}{512^2},\frac{10^3}{512^2}\}$. Using $w_{conv1\_2}$ and $w_{conv5\_2}$ as content layers resulted the style difference appear in random places. Moreover, results using $w_{conv2\_2}$ and $w_{conv3\_2}$ has too firm of a style difference. On the other hand, using content layer $w_{conv4\_2}$ resulted in not too firm or not random style difference. In Fig.~\ref{fig:weights}, we experimented on individual style layers while fixing the content layer to $w_{conv4\_2}$ with weight of $w_{conv4\_2}=10^4$. Each of the results show not incomplete but not overlapping style difference on the content image. Thus, we used all of the style layers to capture complete style difference of the style images.

\begin{figure}[!t]
    \centering
    \includegraphics[width=\columnwidth,trim=0 720 0 0,clip]{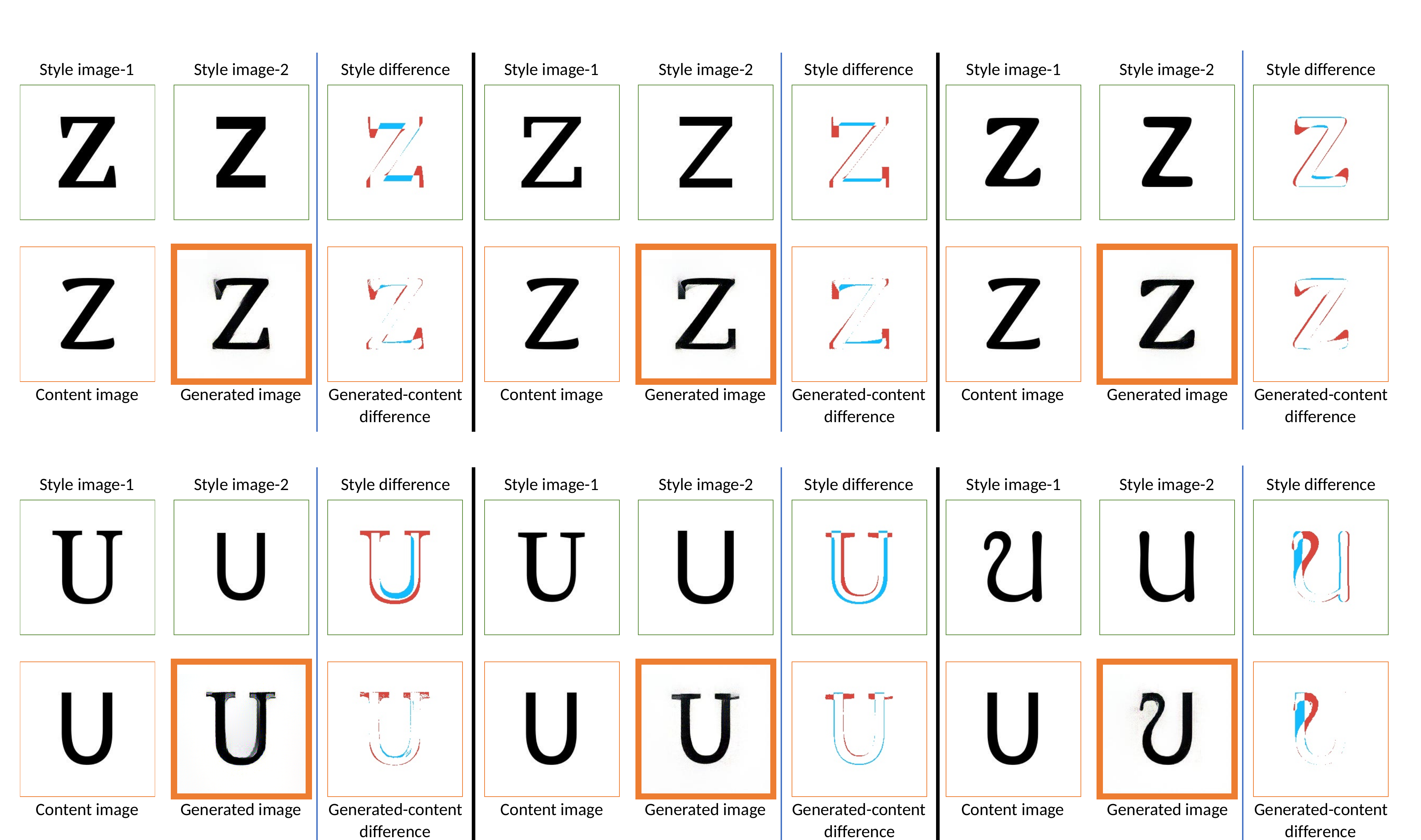}
    \caption{Examples of generating serifs on the content font using the difference between serif font and sans-serif font from the same font family.}
    \label{fig:serif_gen}
\end{figure}

\subsection{Complementing the Content Font}
\subsubsection{Transferring Missing Parts.}
\label{sec:missing_parts}

We experimented on transferring missing parts of a font. As visualized in red in style difference images of Fig.~\ref{fig:missing_part}, style image-2 lacks some parts compared to the style image-1. We will try to transfer this difference of parts to the content image. By using the above parameter settings, the proposed method was able to transfer the missing parts onto the content image as shown in the figure. The transferred parts are visualized in red in the "Generated-content difference" images of the Fig.~\ref{fig:missing_part}. The proposed method transfers the style difference while trying to match the style of the content font. Thus, the most appended part of the content image is connected to the content font. The best results were achieved when the missing parts are relatively small or narrower than the content fonts. Using wider fonts works most of the time. However, the proposed method struggled to style transfer when the difference part is too large or separated. Moreover, missing parts do not only transfer onto the content image, but the style of the missing part is changed to match the style of the content image.

\subsubsection{Generating Serifs.}
\label{sec:serif_gen}

Fig.~\ref{fig:serif_gen} shows the experiments on generating serifs on the content image. Both style images are taken from the same font family. Style image-1 includes serif font, style image-2 includes sans-serif fonts and the content image includes a sans-serif font from different the font family of the style fonts. As shown in the figure, serifs are generated on the content image successfully. Moreover, not only the content font is extended by the serif, parts of it are morphed to include the missing serif as shown in the lower right corner of the figure.

\subsection{Removing Serifs}
\label{sec:removing}

\begin{figure}[!t]
    \centering
    \includegraphics[width=\columnwidth, trim= 390 720 0 0, clip]{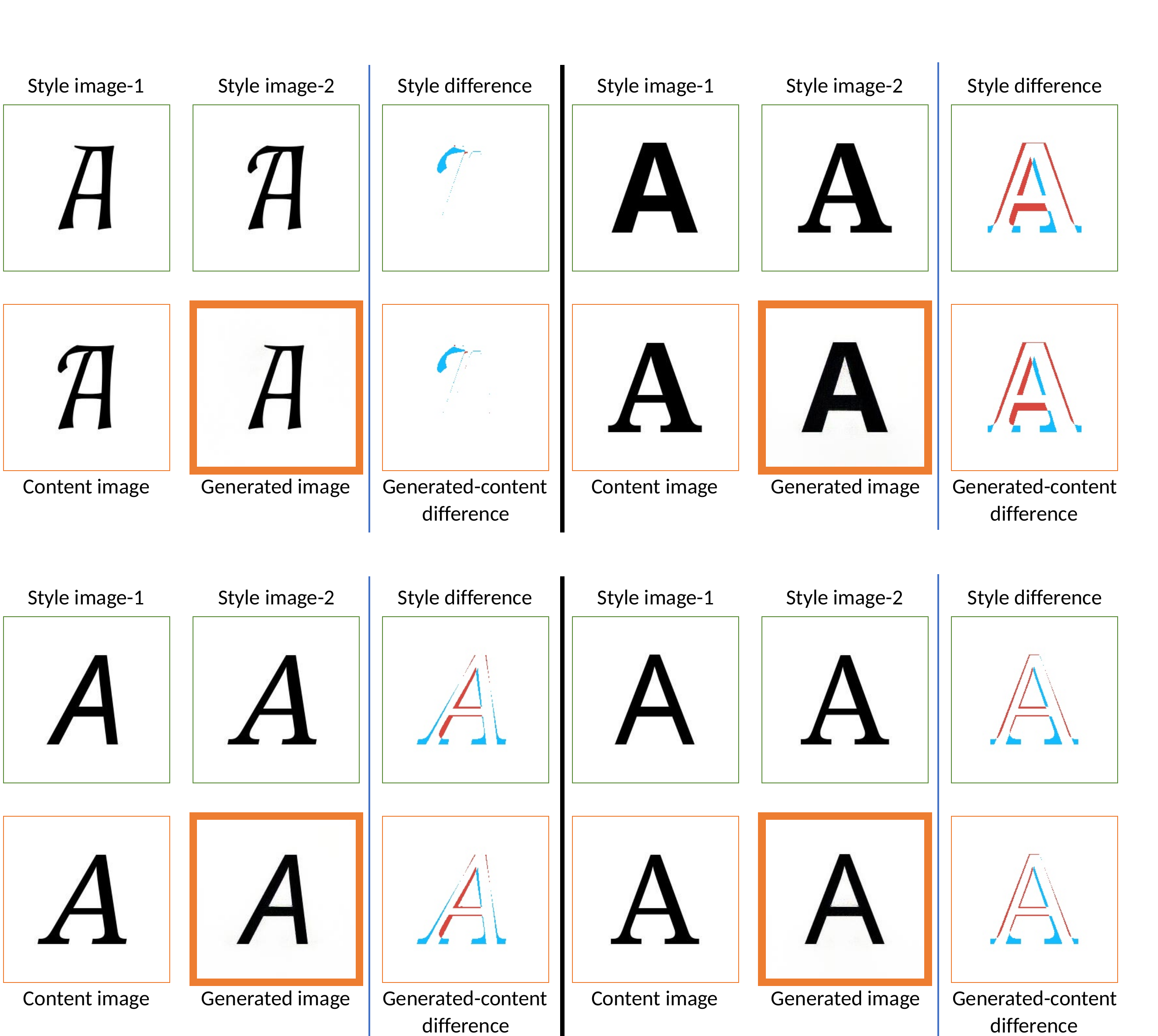}
    \caption{Experiments for removing serifs from the content image by style difference.}
    \label{fig:removing}
\end{figure}

Fig.~\ref{fig:removing} shows the experiments on removing serifs from the content image fonts. Style image-1 includes a font that does not have serif, and style image-2 includes a font that has serifs. By using this difference in serifs, we experimented on removing serifs from the content font. As shown in the figure, serifs have been removed from the content image. However, the font styles of the generated image became different than those of the content image.

\subsection{Transferring Line Width}
\label{sec:bold}

\begin{figure}[t]
    \centering
    \includegraphics[width=\columnwidth, trim=0 720 0 0, clip]{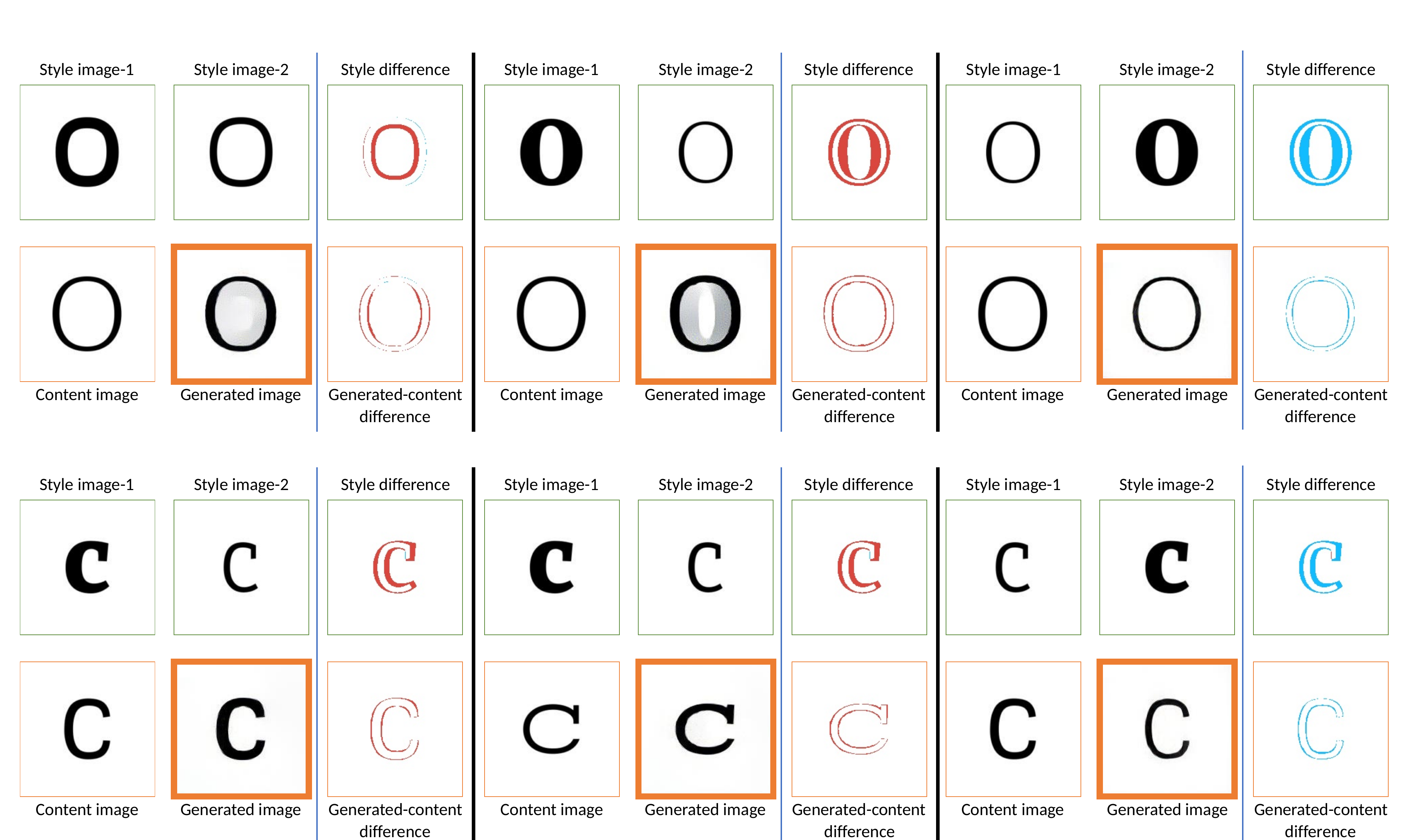}
    \caption{Transferring font line width difference. The first two columns show experiments for widening the font line width of the content font. The third column shows eroding the content font by style difference.}
    \label{fig:bold}
\end{figure}

Fig.~\ref{fig:bold} shows transferring difference of font line width between the style images to the content image to change the font line width from narrow to wide or from wide to narrow. The proposed method was able to change the content font with a narrow line to a font that has a wider line in most cases and vice-versa. However, it struggled to change the wide font line to a narrow font line, when the content font line is wider than the font line in the style image-2.

\begin{figure}[!t]
    \centering
    \includegraphics[width=\columnwidth, trim= 0 50 0 0, clip]{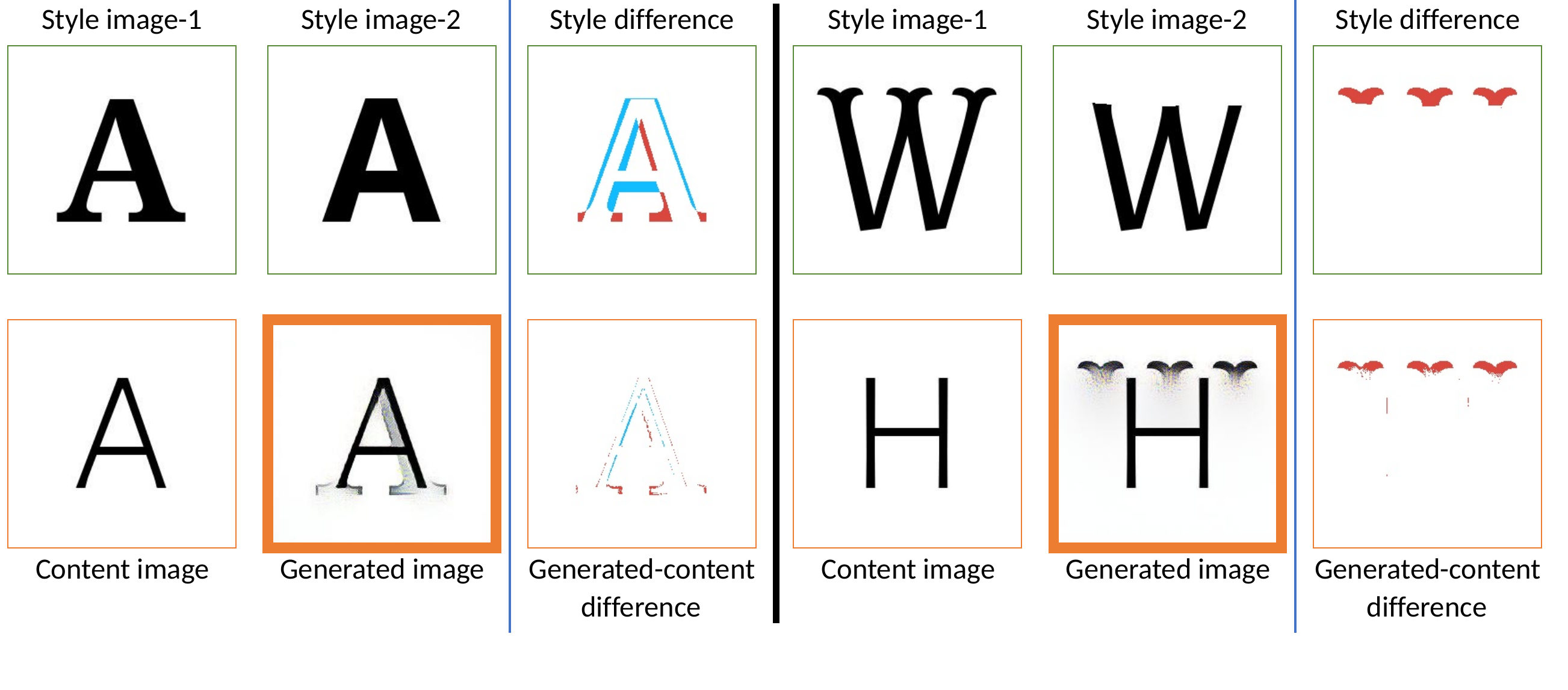}
    \caption{Failure cases. The first experiment shows a failure when styles of the content image and style image-2 are not similar. The second experiment shows a failure case when the input characters are different.}
    \label{fig:failures}
\end{figure}

\subsection{Failure Cases}
\label{sec:failures}

Fig.~\ref{fig:failures} shows failure cases for font generation. On the left side of Fig.~\ref{fig:failures}, fonts from the content image and style image-2 do not have a similar font style. The font in the style image-2 has wide lines, whereas, the font in the content image as narrow lines. The style difference between the style fonts is serif in the foot of the font. Consequently, the proposed method tried to transfer both wideness and serifs, so it resulted in an incomplete font. Moreover, the right side of Fig.~\ref{fig:failures} shows input images that have different characters. Although the font style match between content image and style image-2, the content image is not suitable to receive the font difference between the style images.

\section{Conclusion and Discussion}
\label{sec:conclusion}

In this paper, we introduced the idea of transferring the style difference between two fonts to another font. Using the proposed neural font style difference, we showed that it is possible to transfer the differences between styles to create new fonts. Moreover, we showed that style difference transfer can be used for both the complementing and erasing from the font in the content image with experimental results. However, the input font images must be chosen carefully in order to achieve plausible results. Due to the simple subtraction operation in the content and style difference calculation, font styles of the content image must be similar to font styles of style image-2. So, the font styles of the generated image will become similar to those of style image-1. Another limitation of the proposed method is the processing time due to the back-propagation in each stylization step. These issues can be solved by utilizing the encoder-decoder style transfer method~\cite{Johnson} or adversarial method~\cite{Azadi_2018_CVPR}. However, these methods will have to be trained for the style transfer process first is contrary to the proposed method.

\section*{Acknowledgment} This work was supported by JSPS KAKENHI Grant Number JP17H06100.


\end{document}